\documentclass[sigconf]{acmart}
\pdfoutput=1
\AtBeginDocument{%
  \providecommand\BibTeX{{%
    \normalfont B\kern-0.5em{\scshape i\kern-0.25em b}\kern-0.8em\TeX}}}
    
\newcommand{\sym}[1]{{#1}} 
\usepackage{multirow}

\copyrightyear{2020}
\acmYear{2020}
\setcopyright{acmcopyright}
\acmConference[SIGIR '20]{Proceedings of the 43rd International ACM SIGIR Conference on Research and Development in Information Retrieval}{July 25--30, 2020}{Virtual Event, China}
\acmBooktitle{Proceedings of the 43rd International ACM SIGIR Conference on Research and Development in Information Retrieval (SIGIR '20), July 25--30, 2020, Virtual Event, China}
\acmPrice{15.00}
\acmDOI{10.1145/3397271.3401041}
\acmISBN{978-1-4503-8016-4/20/07}

\begin{document}
\fancyhead{}

\title{Studying Product Competition Using Representation Learning}

\author{Fanglin Chen}
\email{fchen@stern.nyu.edu}
\author{Xiao Liu}
\email{xliu@stern.nyu.edu}
\affiliation{%
  \institution{New York University}
  \city{New York}
  \state{NY}
  \postcode{10012}
}

\author{Davide Proserpio}
\email{proserpi@marshall.usc.edu}
\author{Isamar Troncoso}
\email{itroncos@marshall.usc.edu}
\affiliation{%
  \institution{University of Southern California}
  \city{Los Angeles}
  \state{CA}
  \postcode{90089}
}

\author{Feiyu Xiong}
\affiliation{\institution{Alibaba Group}}
\email{feiyu.xfy@alibaba-inc.com}

\renewcommand{\shortauthors}{Chen, et al.}

\begin{abstract}
Studying competition and market structure at the product level instead of brand level can provide firms with insights on cannibalization and product line optimization. However, it is computationally challenging to analyze product-level competition for the millions of products available on e-commerce platforms. We introduce Product2Vec, a method based on the representation learning algorithm Word2Vec, to study product-level competition, when the number of products is large. The proposed model takes shopping baskets as inputs and, for every product, generates a low-dimensional embedding that preserves important product information. In order for the product embeddings to be useful for firm strategic decision making, we leverage economic theories and causal inference to propose two modifications to Word2Vec. First of all, we create two measures, complementarity and exchangeability, that allow us to determine whether product pairs are complements or substitutes. Second, we combine these vectors with random utility-based choice models to forecast demand. To accurately estimate price elasticities, i.e., how demand responds to changes in price, we modify Word2Vec by removing the influence of price from the product vectors. We show that, compared with state-of-the-art models, our approach is faster, and can produce more accurate demand forecasts and price elasticities.
\end{abstract}



\maketitle

\section{INTRODUCTION}\label{sec:intro}
Identifying key competitors is essential to firms' competitive strategies, such as pricing, product design, and positioning~\cite{desarbo1993non, urban1984testing, bergen2002competitor}. 

However, competition does not occur only among different brands. In mature markets, from high-technology products to grocery items, firms often have long product lines with numerous products under the same brand. As a result, competition also occurs among different products \textit{within} the same brand. Conducting market structure analysis at the product level can provide firms with insights about cannibalization and product line optimization.

In this paper, we propose a machine learning solution to study competition and market structure when the number of products is large. We do so by employing representation learning techniques~\cite{turian2010word,al2013polyglot,mikolov2013efficient} that allow us to overcome the computational constraints encountered in traditional methods. 
In natural language processing, representation learning algorithms take massive collections of text as input and produce continuous word vectors---also called word embeddings---as output. The word embeddings are designed to capture semantic similarities between words: words that appear in similar contexts in the corpus of text (i.e., words that are surrounded by similar words) will be close to each other in the word vector space. Using the same logic, we treat shopping baskets as sentences and products as words and use representation learning to transform each product into a vector. While using text as input, word embeddings capture semantic similarities, using product as input they capture relationships among products. We show that products that share common shopping contexts (i.e., those bought with similar other products), and therefore that are close in the vector space, are more likely to be either complements or substitutes than those far apart in the vector space. 
We use the vectors generated by our algorithm to define two measures, complementarity and exchangeability, that allow us to distinguish between products that are complements and substitutes, respectively.

Because word embeddings techniques can be easily applied to large datasets, our model can be trained on a large number of shopping baskets and products across different categories. This, in turn, allows us to analyze competition among hundreds of thousands of products within just a few hours, and uncover both inter-brand and intra-brand competitive relationships. Additionally, this method is fully automated and does not require any human input or ex-ante assumptions about market segments or structure.

We verify the validity of our complementarity and exchangeability measures by choosing a focal product and identifying the top-3 complement products (those with the highest value for complementarity) and substitute products (those with the highest value for exchangeability). The products obtained show that these measures successfully identify both complements and substitutes.

Combining the product vectors produced by our method with choice models, we are able to estimate and predict consumer choices faster and more precisely. We show that our approach takes 94\% less time than traditional choice models with product fixed-effects or observable product attributes, and increase the out-of-sample hit rates by 4\% (from 14.2\% to 14.8\%). Additionally, when compared with recently developed methods to predict consumer purchases~\cite{ruiz2019shopper}, we achieve higher out-of-sample hit rates (14.1\% vs. 12.8\%).

The remainder of the paper is organized as follows. Section~\ref{sec:related} discusses the related literature and Section~\ref{sec:model} describes our model. Section~\ref{sec:emp} presents empirical applications of our model and results. Section~\ref{sec:conclusions} discusses the implications of our findings and provides future research directions and concluding remarks.

\section{RELATED LITERATURE}\label{sec:related}

Our paper is based on the stream of representation learning literature. The basic model that we build on uses the representation learning algorithm, specifically those from the family of models based on neural networks in the neural language models literature. There are several approaches to learn word embeddings using different models and objective functions. For example, ~\cite{bengio2003neural} propose a model whose objective is to predict the next word in a sequence based on the previous words. In~\cite{collobert2011natural}, the goal is to capture the entire context of words based on the whole sentences. Finally,~\cite{mikolov2013efficient} predict the surrounding words (previous and next words) based on the current word.
Our method uses the Word2Vec model by~\cite{mikolov2013efficient}, who propose a neural network model with a single-layer architecture to learn the distributed representations of words. The authors also introduce a \textit{negative sampling} procedure, an approximation to the original objective function that not only improves the quality of the embeddings but also considerably increases the speed of training. This procedure distinguishes the context word (i.e., word to predict) from random draws from the corpus by encouraging the current word and the context (random) word to have a larger (smaller) co-occurrence probability~\cite{mnih2013learning}. 

Only recently has the marketing literature begun to take advantage of machine learning techniques such as representation learning. For example,~\cite{gabel2019p2v} employ representation learning to transform products into vectors using information from shopping baskets.~\cite{gabel2019p2v} derive latent product attributes and uncover market structure by reducing the dimensionality of latent properties. Differently from ours, their main goal is to propose a scalable method to understand the market structure of all products in a retailer's assortment. Our goal is to measure how sensitive consumers are to price changes (i.e., price elasticities) and perform demand forecasts by combining representation learning and choice models. Therefore, in addition to creating product positioning maps, we deal with issues such as controlling the effect of price on product co-occurrences, addressing price endogeneity, and distinguishing complements from substitutes.

To date, the closest paper to ours comes from a trio of computer scientists and economists who propose a hierarchical latent variable model of market baskets, called SHOPPER~\cite{ruiz2019shopper}.
SHOPPER aims to predict the items in the ordered shopping baskets sequentially with the latent variables, including item dependencies, customer preferences, price sensitivities, and seasonal effects. The authors focus on prediction and obtain impressive results on a huge dataset of consumer purchases. However, this comes at the cost of a complex model that requires the estimation of a large number of parameters, and does not take into consideration price endogeneity.

Differently from SHOPPER, we provide a model that can achieve similar prediction accuracy, but that relies on a much smaller number of parameters, a shorter estimation time, and provides unbiased estimations of price elasticities. Moreover, our focus is the marketing discipline, and thus we provide valuable marketing insights that are not discussed in~\cite{ruiz2019shopper}.

\section{MODEL}\label{sec:model}

In this section, we describe the model to conduct product-level competition analysis with a large number of products. 

In traditional choice models, each product is usually represented as a fixed-effect parameter, thus assuming that products are independent of each other. In doing so, when looking at large product categories with hundreds or thousands of products, these models have to estimate a large number of parameters, which often translates to extremely high computational costs. In practice, however, consumers consider a limited number of product characteristics (e.g., taste, aesthetics, or package size, to mention a few) when deciding what to purchase~\cite{hauser2014consideration}. This means that, in practice, it would be enough to represent products using these few attributes to fully capture the differences across products and, therefore, consumer preferences. One way to achieve this would be to \textit{manually} choose the product attributes that matter and then estimate the choice model. However, it is difficult to determine the number of attributes and which attributes to use, because they are often context-dependent (e.g., the relevant attributes for coffee are likely different from those for household cleaners). Moreover, even if it were possible to choose the attributes by category, some product attributes may be hard to observe and, therefore, measure (e.g., the appeal of packaging).

To overcome these limitations, we propose an unsupervised method to learn (latent) product attributes by leveraging representation learning. We use representation learning algorithms to learn product embeddings that allow us to (i) obtain a meaningful product representation in a lower-dimensional space, and (ii) uncover relationships (complements, substitutes) among products. We introduce Product2Vec, which learns the vector representations of products from product co-occurrences in shopping baskets. These product vectors are then incorporated into a choice model to estimate price elasticities as well as to predict consumer purchases.

\subsection{Product2Vec: Learning the Vector Representations of Products}\label{sec:Product2Vec}

We adopt the framework of Word2Vec~\cite{mikolov2013efficient} to create our Product2Vec model. Product2Vec can transform products into low-dimensional vectors with continuous elements, using the information regarding the composition of shopping baskets. The objective function used in the training is:
\begin{equation}
\label{eq:p2v_objective}
    \mathcal{V, V'}= \mathop{\arg\max}_{\mathcal{V, V'}} \ \sum\limits_{b \in B}\sum\limits_{s_i \in b}\sum\limits_{-c\le j \le c} \text{log}\ \mathbb{P}(s_{i+j}|s_i; \mathcal{V, V'}) 
\end{equation}
In Equation~\ref{eq:p2v_objective}, $\mathcal{V, V'}$ are the sets of ``input'' and ``output'' product vectors.
Specifically, let $v_s, v_s^{'} \in \Re^M$ be the $M$-dimensional ``input'' and ``output'' vectors of product $s$,  and $S$ be the total number of unique products. Then $\mathcal{V}=\{v_s\}_{s=1}^S$, $\mathcal{V'}=\{v_s^{'}\}_{s=1}^S$ are the collections of $v_s$ and $v_s^{'}$ respectively. $B$ is the set of all shopping baskets, $b$ is a shopping basket in the set $B$, $s$ is an product in the basket, $c$ is the length of the context for product sequences, and $\mathbb{P}(s_{i+j}|s_i)$ is the conditional probability of observing the context product $s_{i+j}$ given the focal product $s_i$.\footnote{If multiple units of an product are purchased, we treat it as if the product only appears once. We do not consider purchase quantity because it has little to do with relationships among products.\label{fn3}} Following \cite{mikolov2013distributed}, we define the conditional probability using the softmax function:
\begin{equation}
\label{eq:p2v_softmax}
    \mathbb{P}(s_{i+j}|s_i; \mathcal{V, V'}) = \frac{exp(v_{s_i}^T\ v_{s_{i+j}}^{'})}{\sum_{s=1}^{S}exp(v_{s_i}^T\ v_s^{'})}
\end{equation}

Because the cost of computing the denominator is proportional to the total number of unique products ($S$), it is impractical to directly calculate $\mathbb{P}(s_{i+j}|s_i; \mathcal{V, V'})$. Instead, we employ the negative sampling technique to approximate the log probability of the softmax~\cite{mikolov2013distributed}:
\begin{equation}
\label{eq:p2v_negative_sampling}
    \text{log}\ \mathbb{P}(s_{i+j}|s_i; \mathcal{V, V'}) = \text{log}\ \sigma(v_{s_i}^T\ v_{s_{i+j}}^{'}) + \sum\limits_{k=1}^K \mathbb{E}_{s_k} \text{log}\ \sigma(-v_{s_i}^T\ v_{s_k}^{'}).
\end{equation}
In Equation~\ref{eq:p2v_negative_sampling}, $s_k$ is an product randomly drawn from the whole training set based on the distribution of purchase frequency, $K$ is the number of negative samples for each focal product, and $\sigma()$ is the sigmoid function. Equation~\ref{eq:p2v_negative_sampling} consists of two components: the first component maximizes the probability that the current product occurs together with its context products, while the second minimizes the likelihood that the current product appears along with some randomly selected, irrelevant products. In other words, this objective function distinguishes observations from noise. 

\subsection{Distinguishing Between Substitutes and Complements} \label{sec:substitutes_and_complements}

An important issue we need to deal with is the fact that product embeddings does not directly reveal whether two products are complements or substitutes.\footnote{Complements and substitutes are defined with respect to the whole set of products, not restrained to products in the same basket.} As we discussed above, vector similarity suggests that two products are likely to be \textit{related}, and this relationship can be either one of complements or substitutes. To resolve this issue, we follow the approach adopted by~\cite{ruiz2019shopper}, and define two measures, \textit{complementarity} and \textit{exchangeability}, to identify complements and substitutes respectively.

Roughly speaking, two products are likely to be complements if the interaction between their latent dimensions is positive, and two products are likely to be substitutes if they predict similar purchase patterns for the rest of the products in the store. We provide the formal definitions of complementarity and exchangeability next.

\paragraph{Complementarity.} 
We consider two products A and B to be complements if the conditional probability of purchasing A (or B) given B (or A) being already in the basket is high. In simple words, two products are complements if they are very likely to be purchased together. Formally, we compute the complementarity between products A and B as:
\begin{align}
\begin{split}
\label{eq:complements}
C_{AB} &= \frac{1}{2} \Big[ P(A|B) + P(B|A) \Big]\\
& \propto \frac{1}{2} \left( v_A^T \cdot v_B' + v_B^T \cdot v_A' \right)
\end{split}
\end{align}
where v and v' are the input and output vectors in Equation~\ref{eq:p2v_softmax}.

\paragraph{Exchangeability.} 
We consider two products A and B to be exchangeable if they induce similar distributions of the conditional purchase probability of the rest of the products in the store. To put it simply, A and B are exchangeable if they interact similarly with other products. Formally, we measure the similarity between two distributions using the negative KL divergence~\cite{kullback1951information} and compute the exchangeability between products A and B as:
\begin{align}
\begin{split}
\label{eq:exchangeables}
E_{AB} &= -\frac{1}{2} \Big[ KL(p(\cdot|A) \parallel p(\cdot|B)) + KL(p(\cdot|B) \parallel p(\cdot|A)) \Big] \\
&= -\frac{1}{2} \sum_{k \neq A, B} \Bigg[ p(k|A) \cdot \log \left( \frac{p(k|A)}{p(k|B)} \right)  + p(k|B) \cdot \log \left( \frac{p(k|B)}{p(k|A)} \right)  \Bigg]
\end{split}
\end{align}
where $p(\cdot|A), p(\cdot|B)$ are the distributions of the conditional purchase probability of all the other products in the store given $A$ or $B$ is purchased, and the KL divergence between the two distributions is further calculated using the conditional purchase probability of any product $k$ given $A$ or $B$, $p(k|A), p(k|B)$. Based on Equation~\ref{eq:p2v_softmax}, we can represent $p(k|A), p(k|B)$ with the corresponding product vectors.

If two products have a high exchangeability score, they are likely to be substitutes.\footnote{A low value of the KL divergence indicates that the two probability distributions being compared are very similar. Therefore, a high exchangeability score indicates that $p(\cdot|A)$ and $p(\cdot|B)$ are similar, i.e., products A and B are likely to be substitutes.} However, as discussed in~\cite{ruiz2019shopper}, two products that are frequently purchased together (i.e., two products that are complements) will also have a high exchangeability score. This is because complements tend to appear in the same basket and, therefore, they have similar interactions with other products. To exclude possible complements, we define two products as substitutes if their exchangeability score is high and their complementarity score is low.

\subsection{Choice Model} \label{sec: choice model}

After having transformed products into low-dimensional vectors using purchase information from all product categories, we focus on competitive analysis among products within one specific category. The starting point of this analysis is the well-known logit model of consumer choices~\cite{guadagni1983logit}. For each consumer $i\in\{1, ..., I\}$ and each time period $t\in\{1, ..., T\}$, we describe a product $j\in\{1, ..., J\}$ by a product-specific dummy variable $\alpha_j$, price $P_{jt}$ and other marketing variables $\vec X_{jt}$, such as whether the product is under a promotion and/or advertisement. The utility of product $j$ for consumer $i$ at time period $t$ is:
\begin{equation}
\label{eq:choice_model}
    U_{ijt} = \alpha_j + \beta P_{jt} + \vec X^{'}_{jt} \vec \gamma + \epsilon_{ijt} 
\end{equation}
where $\alpha_j$ captures consumers' intrinsic preferences of unobserved product characteristics, $\beta$ and $\vec \gamma$ measure consumers' sensitivities to price and other marketing variables, and $\epsilon_{ijt}$ is a type I extreme value error term.

In the choice model, the utility $U_{ijt}$ is further used to calculate the choice probability using the softmax function, so a higher utility $U_{ijt}$ is associated with a higher choice probability of product $j$. In Section~\ref{sec:model performance}, we use the choice model to predict consumer choices. Given the estimated parameters, we predict that the product with the highest utility among all products is purchased, and calculate the hit rate by comparing our predictions with actual purchases.

When the number of products is large, the number of coefficients to estimate in Equation~\ref{eq:choice_model} is also large (recall that there is one dummy variable for each product), which may make the estimation very slow or, in some cases, intractable. 
To reduce the parameters to be estimated, we replace product dummies $\alpha_j$ with product vectors $v_j$. The model specification becomes:
\begin{equation}
\label{eq:homogeneous_choice_model}
    U_{ijt} = \sum_{m=1}^{M}\alpha_m v_{j m} + \beta P_{jt} + \vec X^{'}_{jt} \vec \gamma + \epsilon_{ijt} 
\end{equation}
where $v_{j m}$ is the $m$th dimension of the $M$-dimensional product vector $v_j$, and $\alpha_m$ is the model parameter corresponding to the $m$th dimension. Since $M<J$, using the latent representations of products decreases the number of parameters to estimate.

Finally, to account for consumer heterogeneity, we follow the mixed logit model~\cite{mcfadden2000mixed} and set the price coefficients to be consumer-specific and normally distributed across the population:
\begin{equation}
\label{eq:heterogeneous_choice_model}
    U_{ijt} = \sum_{m=1}^{M}\alpha_m v_{j m} + \beta_i P_{jt} + \vec X^{'}_{jt} \vec \gamma + \epsilon_{ijt}  
\end{equation}
where $\beta_i \sim \mathcal{N}(\bar{\beta}, \sigma_{\beta}^{2})$.

\subsection{Addressing Price Issues} \label{sec: addressing price issues}
The specifications in Equations~\ref{eq:homogeneous_choice_model} and \ref{eq:heterogeneous_choice_model} may suffer from two issues related to price. The first issue is that in choice models, price is likely to be endogenous. For example, unobserved demand shocks could potentially drive both prices and consumer choices, and directly regressing consumers choices on prices will overlook such shocks and result in biased estimates. As it is customary in the economics literature, we deal with this problem by using an instrument variable~\cite{berry1995automobile}. 

The second issue is related to the way in which we create the product vectors using representation learning. As we discussed in Section~\ref{sec:model}, vectors are learned from product co-occurrences. Co-occurrences can be influenced by observed and latent product attributes as well as by price similarity. For example, consumers are more likely to buy and substitute between Aquafina and Dasani water rather than VOSS water because the first two products have similar prices, while the last one is much more expensive. If vector similarities are mostly driven by product prices, then one dimension of the product vectors could already incorporate price information; this could bias price elasticities and make demand predictions less accurate. To avoid this problem, we propose two different solutions that we describe in detail in Section~\ref{sec:dealwithprice}.

\subsubsection{Price Endogeneity}

To deal with price endogeneity, we use the control function approach~\cite{petrin2010control}. This approach consists of two steps. In the first step, the endogenous variable (i.e., price) is regressed on the instrument variable. We use the average price for the same product and week in other stores of the same chain as the instrument variable~\cite{berry1995automobile}.
The residuals of the first stage, which represent the component correlated with unobserved pricing shocks, are then retained. In the second step, the choice model is estimated with the retained residuals as an additional variable. Formally, the first stage is as follows:
\begin{equation}
\label{eq:iv_first_stage}
    P_{jt} = \sum_{m=1}^{M}\tau_m v_{j m} + \tau_Z Z_{jt} +  \vec X^{'}_{jt} \vec \tau_X + \eta_{jt}
\end{equation}
where $Z_{jt}$ is the instrument discussed above, and $\tau_m, \tau_Z, \vec \tau_X$ are the coefficients used to model the price pattern. We then estimate the residuals of the first stage, $\hat{\eta}_{jt}$, and include them in the second step of the control function approach:
\begin{equation}
\label{eq:iv_second_stage}
    U_{ijt} = \sum_{m=1}^{M}\alpha_m v_{j m} + \beta_i P_{jt} + \vec X^{'}_{jt} \vec \gamma + \delta \hat{\eta}_{jt} + \epsilon_{ijt} 
\end{equation}

\subsubsection{Revised Product2Vec}\label{sec:dealwithprice}

We remove the influence of price on the product vectors by incorporating price into our Product2Vec model explicitly. We do so by treating price as an additional dimension in the product vectors. Using the original Word2Vec model, two products will have similar product vectors if they are purchased in similar baskets. However, observed similar baskets could be driven by two forces: two products may have similar product attributes or prices. By representing price with an extra dimension, we could separate product attributes from prices, and get product vectors that are not influenced by price.

In the training process, we replace the product vector $v_s$ with $\bigl( \begin{smallmatrix} v_s \\ P_s \end{smallmatrix} \bigr)$, where $P_s$ is the price of product $s$.\footnote{To control for the magnitude of prices, we first normalize them within each category and then use them as input for the model.} Then, in each iteration of the optimization process, instead of updating the full vectors of $M+1$ dimensions, we only update the first $M$ values, leaving the last dimension ($P_s$) unchanged (i.e., the value observed in the data). After training $\bigl( \begin{smallmatrix} v_s \\ P_s \end{smallmatrix} \bigr)$, we drop $P_s$ and keep only $v_s$ for further analysis, which is a vector free of the price influence.

Based on the mechanism of the Product2Vec model, if two products $C$ and $D$ appear in similar contexts, their product vectors $\bigl( \begin{smallmatrix} v_C \\ P_C \end{smallmatrix} \bigr)$ and $\bigl( \begin{smallmatrix} v_D \\ P_D \end{smallmatrix} \bigr)$ should be close to each other, and their inner product should be large. A large inner product could be driven by both price and non-price dimensions. Assume that C and D have very different prices. In this case, the inner product of $P_C$ and $P_D$ is small; therefore, the inner product of $v_C$ and $v_D$  must be large. However, if C and D have quite similar prices, then the inner product of $P_C$ and $P_D$ is large, and therefore the inner product of $v_C$ and $v_D$ are not necessarily large.

\subsection{Incorporating Complementarity and Exchangeability}

As we discussed in Section~\ref{sec:substitutes_and_complements}, the complementarity and exchangeability measures allow us to distinguish between pairs of products that are complements or substitutes, thus providing valuable information about the relationships among products.
If one product is complementary to the other products in the basket, then the utility of choosing this product will increase. On the contrary, if one product substitutes the existing products, then it has a smaller chance of being purchased. Therefore, in addition to the product vectors, we incorporate these two new variables in the choice model. Building upon Equations~\ref{eq:iv_first_stage} and \ref{eq:iv_second_stage}, our final choice model becomes:
\begin{equation}
\label{eq:iv_first_stage_w/int}
    P_{jt} = \sum_{m=1}^{M}\tau_m v_{j m} + \tau_Z Z_{jt} + \vec X^{'}_{jt} \vec \tau_X + \tau_C C_{j b} + \tau_E E_{j b} + \eta_{jt}
\end{equation}

\begin{equation}
\label{eq:iv_second_stage_w/int}
    U_{ijt} = \sum_{m=1}^{M}\alpha_m v_{j m} + \beta_i P_{jt} + \vec X^{'}_{jt} \vec \gamma + \lambda C_{j b} + \mu E_{j b} + \delta \hat{\eta}_{jt} + \epsilon_{ijt} 
\end{equation}
where $C_{j b}$ and $E_{j b}$ are the average complementarity and exchangeability scores between product $j$ and each of the other products in the basket $b$.

\section{RESULTS}\label{sec:emp}
\subsection{Data}
We test our model using the IRI's scanner panel data~\cite{bronnenberg2008database}, which contains transaction information of 30 product categories and 13,124 unique products purchased by 5,214 households across 53 stores at the weekly level for 52 weeks. The total number of shopping baskets amounts to 280,052. We divide this data into three sets: 40\% training set, 40\% estimation set, and 20\% test set.
we set the number of dimensions of the product vectors $M=20$, the length of context $c=5$, and the number of negative samples $K=5$.\footnote{We experiment with $M=20, 50, 100, c=5, 10, 15, 20, K=5, 10, 15, 20$, and the results are similar.}

\subsection{Model Validation}\label{sec:model validation}

We assess the capability of our complementarity and exchangeability metrics (Equations~\ref{eq:complements} and~\ref{eq:exchangeables}, respectively) by choosing a focal product and computing these metrics for all the remaining products in the sample. We expect that complements are the products with the highest complementarity scores, and substitutes are the products with the highest exchangeability scores, conditional on having low complementarity scores.

We report the results for three examples in Table~\ref{tab:complements_substitutes_iri_data}. For each focal product (column 1), we list the top-3 products with the highest complementarity scores (column 2) and the top-3 products with the highest exchangeability scores and low complementarity scores (column 6), including the values for complementarity and exchangeability for each product pair. We observe that our approach recovers reasonable complements pairs such as cereal and coffee, beer and cigarettes, beer and frozen pizza, and preferences for private label products. The same is true for substitutes: the model successfully recovers pairs of competing products from the same brand (e.g., two Kelloggs cereals or two private label oats) or different brands (e.g., beers from different brands). 

\begin{table*}
\centering
\caption{Examples of Complements and Substitutes Found by Revised Product2Vec}
\label{tab:complements_substitutes_iri_data}
\begin{tabular}{llcclcc}
\toprule
  & \multicolumn{3}{c}{Highest Complementarity}                 
  & \multicolumn{3}{c}{Highest Exchangeability}    \\
 \cmidrule(r){2-4} \cmidrule(l){5-7}
Focal Product & Product Name & Comp. & Ex. & Product Name & Comp. & Ex. \\
\midrule
 & \begin{tabular}{@{}l@{}}NESTLE NESQUIK \\ MILKSHAKE 0.8438PT\end{tabular}              & 1.419           & -0.883           & \begin{tabular}{@{}l@{}}KELLOGGS FROSTED \\ FLAKES REGULAR 20OZ\end{tabular}      & 0.175           & -0.433              \\
\begin{tabular}{@{}l@{}}KELLOGGS RICE  \\ KRISPIES TREATS 14.2OZ \end{tabular}        & \begin{tabular}{@{}l@{}}HILLS BROTHERS \\ GROUND COFFEE 11.5OZ\end{tabular}         & 1.337           & -0.911              & \begin{tabular}{@{}l@{}}KELLOGGS RAISIN \\ BRAN REGULAR 20OZ\end{tabular}       & 0.343           & -0.463           \\
 &\begin{tabular}{@{}l@{}}EIGHT O CLOCK WHOLE \\ BEAN COFFEE 12OZ \end{tabular}       & 1.292           & -1.697              & \begin{tabular}{@{}l@{}}KELLOGGS FROSTED MINI \\ WHEATS REGULAR 19OZ \end{tabular} & 0.227           & -0.484           \\ 
\midrule
& \begin{tabular}{@{}l@{}}PRIVATE LABEL GROUND \\ DECAF COFFEE 26OZ  \end{tabular}       & 1.948           & -0.392           &  \begin{tabular}{@{}l@{}}PRIVATE LABEL HNYNT \\ OAT BOX NUT 14OZ \end{tabular}       & 0.798           & -0.335           \\
 \begin{tabular}{@{}l@{}}PRIVATE LABEL \\ OATS BOX 15OZ  \end{tabular}                                & \begin{tabular}{@{}l@{}}PRIVATE LABEL \\ CONDENSED WET SOUP \end{tabular}      & 1.656           & -0.683               & \begin{tabular}{@{}l@{}}POST BRAN FLAKES \\ CEREAL BOX 16OZ \end{tabular}           & 0.335           & -0.344           \\
 & \begin{tabular}{@{}l@{}}WELLS BLUE BUNNY LITE \\ NONFAT YOGURT 1.5PT \end{tabular} & 1.595           & -0.730               & \begin{tabular}{@{}l@{}}HORMEL WRANGLERS \\ FRANKFURTERS 8CT 16OZ \end{tabular}   & 0.486           & -0.350             \\ 
\midrule
 & \begin{tabular}{@{}l@{}}NOW ULTRA LIGHTS \\ CIGARETTES 1 COUNT \end{tabular}         & 1.079           & -0.958                 & \begin{tabular}{@{}l@{}}POINT HONEY \\ LIGHT BEER 72OZ \end{tabular}             & -0.238          & -0.191             \\ 
  \begin{tabular}{@{}l@{}}HACKER PSCHORR \\ WEISSE BEER 144OZ \end{tabular}    & \begin{tabular}{@{}l@{}}TOMBSTONE HARVEST\\ FROZEN PIZZA 17.4OZ \end{tabular}    & 0.896           & -0.596              & \begin{tabular}{@{}l@{}}SPATEN OKTOBERFEST \\ BEER 144OZ     \end{tabular}          & -0.631          & -0.230          \\
   & \begin{tabular}{@{}l@{}} TOMBSTONE FROZEN \\ PIZZA 17.4OZ     \end{tabular}            & 0.733           & -0.664              & \begin{tabular}{@{}l@{}} BERGHOFF SOLSTICE \\ WHEAT BEER 72OZ    \end{tabular}   & -0.416          & -0.243         \\
\bottomrule
\end{tabular}
\end{table*}

\subsection{Model Performance}\label{sec:model performance}

\subsubsection{Comparison with Traditional Choice Models}

In this section, we compare the performance of our models with state-of-the-art product-level choice models. Specifically, we compare the precision of the price coefficient, model fit, hit rates, and run time with the following models:
\begin{enumerate}
    \item \textbf{Target model}, a choice model that contains a dummy (fixed effect) for every product in the dataset. It is the best possible model because it includes a parameter for every product. However, it is not scalable because the number of parameters grows linearly with the number of products in the dataset. We consider this the benchmark model against which we compare our approach.
    \item \textbf{FH model}~\cite{fader1996modeling},
    a choice model that uses observable product attributes to characterize a large set of products in a parsimonious manner. Despite using fewer parameters than the target model, ~\cite{fader1996modeling} show that it can achieve good results if the attributes chosen are able to characterize consumer choices. 
\end{enumerate}

All the estimations are performed on the same machine, an Ubuntu system with  2 $\times$ 16-core Intel Xeon CPU E5-2697A v4 @ 2.60GHz,  512Gb of RAM, and a GPU NVIDIA Quadro P4000, and using Stata 16 MP (2 cores) for the choice models.\footnote{While Product2Vec can be estimated on any recent personal computer, SHOPPER requires a GPU. Therefore, for comparison purposes we perform all the estimations on the same server.}

We report the estimates in Table~\ref{tab:mixlogit-dim20-model2} for the mixed logit model. In column 1, we report the estimates of the target model; in column 2, we report the estimates of the FH model; in columns 3 and 4, we report the estimates of choice models that use product vectors obtained from Product2Vec, without and with the complementarity and exchangeability measures; in columns 5 and 6, we report the estimates of choice models that use product vectors obtained from Revised Product2Vec, without and with the complementarity and exchangeability measures.

\begin{table*}[t]
\centering
  \caption{Results of the Mixed Logit Model}
  \label{tab:mixlogit-dim20-model2}
  \begin{tabular}{lcccccc}
    \toprule
&\multicolumn{1}{c}{Target}&\multicolumn{1}{c}{FH}&\multicolumn{2}{c}{p2v}&\multicolumn{2}{c}{rp2v}\\
\cmidrule(lr){2-2} \cmidrule(lr){3-3} \cmidrule(lr){4-5}\cmidrule(lr){6-7} 
Log price ($\beta$)        &   -3.145\sym{***}     & -3.942\sym{***} & -2.795\sym{***}&  -2.894\sym{***} & -3.518\sym{***} & -3.445\sym{***} \\
&     (0.272)    & (0.130)      & (0.154)   &     (0.157) & (0.140)        & (0.137)                          \\
Large feature ($\gamma_1$)     &    0.123     & -0.124                           & 0.231\sym{**} & 0.195\sym{**}   & -0.078                     &  -0.062                        \\
                  & (0.096)       & (0.075)                         & (0.071)           &    (0.072)                   & (0.068)                         & (0.066)                         \\
Major display ($\gamma_2$)    &   0.658\sym{***}     & 0.698\sym{***}  & 1.184\sym{***}  &       1.191\sym{***}   & 0.754\sym{***}  &  0.799\sym{***}  \\
                  &     (0.160)      & (0.140)                          & (0.116)        &     (0.119)                                               & (0.119)                          & (0.119)                          \\
Complementarity ($\lambda$) &        &                   &  &       0.726\sym{***}          &                                 &      0.705\sym{***}                             \\
                  &        &                                  & &    (0.049)        &         &   (0.035)                        \\
Exchangeability ($\mu$) &        &  &    &      -0.279                 &                                  &    -0.033   \\
                  &        &                                  &    &     (0.239)                                             &                                  & (0.047)                         \\ 
    \midrule
No. of observations                 &    554434     & 554434                           & 554434  & 554434                             & 554434                           & 554434                           \\

Log likelihood                &    -19712.4    & -20410.3                         & -19964.8   &    -19825.8                                     & -20140.1                         & -19895.5                          \\

AIC               &  39632.8      & 40878.6                          & 39979.5        &     39705.6                                  & 40330.2                          &  39844.9                          \\

BIC               &  40800.2       & 41204.1                          & 40260.2          &     40008.7                      & 40610.9                          &  40148.0                          \\

In-sample hit rate      &  0.136        & 0.115                           & 0.133            &      0.137                     & 0.127                           &  0.144                           \\

Out-of-sample hit rate        &  0.142        & 0.122                           & 0.141              &      0.142                   & 0.119                           & 0.148                           \\

Run time    &   08:52:20     & 00:56:45                            & 00:32:06 & 00:42:05  & 00:32:01    & 00:34:30       \\
    \bottomrule
\multicolumn{6}{l}{Significance levels: * p<0.1, ** p<0.05, *** p<0.01.}
\end{tabular}
\end{table*}

\paragraph{Price coefficient.} We observe that the price estimates in the choice models with product vectors (columns 3 to 6) are closer to the target than the estimates from the FH model. We argue that this is because the vectors capture product characteristics that are not observable in the data. While the FH model controls for as many observable product characteristics as possible, our results suggest that similar or better results can be achieved by controlling for latent product dimensions (i.e., product embeddings). This finding makes our approaches particularly appealing in cases where product characteristics are not readily available or when the nature of the product category requires defining attributes not easily measurable (e.g., while aesthetics or style are important characteristics for clothing, they are often hard to measure and quantify).

\paragraph{Model fit.} We observe that our models using product vectors obtain better fit than the FH model, as the former have larger log likelihood, and smaller AIC and BIC values.\footnote{AIC (Akaike information criterion) and BIC (Bayesian information criterion) are two statistical criteria for model selection, and models with smaller AIC and BIC values are preferred.} Moreover, our models often have a lower BIC value than the target model.

\paragraph{Hit rate.} We look at both the in-sample and out-of-sample hit rates. Compared with the FH model, our models achieve higher in-sample and out-of-sample hit rates in all cases. Our best model is Revised Product2Vec with exchangeability and complementarity (column 5). With a 14.8\% out-of-sample hit rate, it outperforms the target model (14.2\%) by 4\%. 

\paragraph{Run time.} For the mixed logit model, the time saved is even more significant, from around 9 hours for the target model to only half an hour for our models, which is a 94\% reduction in run time. The run time of the FH model is comparable to ours; however, our approaches always outperform it.

Overall, the results discussed above suggest that our models produce results that are generally better than the FH model, and that are always comparable with the target model, although the target model includes more parameters than ours.
Further, these results suggest that taking into account the price influence (i.e., Revised Product2Vec) is important as it allows us to obtain more precise estimates of price elasticities. Ignoring the price influence (columns 3 and 4) leads to estimated price elasticities that are generally smaller than the correct estimate. Also, among our models, Revised Product2Vec with exchangeability and complementarity (column 6) has consistently the highest hit rates in predicting consumer choices.

\subsubsection{Comparison with SHOPPER}

Finally, we compare our model to a recently developed approach that models large-scale market baskets data, SHOPPER~\cite{ruiz2019shopper}. SHOPPER is a hierarchical latent variable model based on sequential choices. It assumes that on each shopping trip, a consumer enters the store and sequentially adds items into her basket conditional on the items added previously. The utility derived from choosing each product is a function of its price, its latent features, its interaction with the latent features of the products that are already in the basket, consumer's unobservable preferences, and seasonality.

Both our approach and SHOPPER use latent representations of products in a discrete choice model setting. However, there are substantial differences between our models and SHOPPER that are worth discussing. The first one is that SHOPPER learns the latent representations of products and estimates choices simultaneously, while our approach is a two-step procedure in which we first estimate the latent representations of products (i.e., the vectors) and then estimate consumer choices by incorporating these vectors as covariates into a choice model. The second difference is that SHOPPER requires significantly more computational resources than our method. SHOPPER incorporates many more variables (e.g., user-specific vectors), and its estimation requires the use of a GPU, while our models can be easily estimated on any personal computer. 
The third difference is that the current version of SHOPPER does not accommodate additional covariates (such as additional marketing mix variables), while our models are quite flexible in terms of accommodating new variables. Finally, SHOPPER does not deal with price endogeneity, while we employ the control function approach to address this problem. 

In order to perform a fair comparison, we re-estimate our choice models using the same set of covariates to those included in SHOPPER. Hence, we drop the \textit{feature} and \textit{display} variables from the choice models, keeping only price and product-specific vectors. We also estimate the target model, in which the product-specific vectors are product fixed effects, and the FH model, in which the product-specific vectors are observable product attributes such as brand, size, additives, and packaging.

We report the results in Table~\ref{tab:shopper-comparison-iv}, where we compare SHOPPER and the choice models in terms of price elasticities and in and out of sample hit rates (i.e., hit rates in the training and test sample, respectively). For each model, the first column reports the results without the complementarity and exchangeability measures, and the second column with complementarity and exchangeability measures.\footnote{The run times of SHOPPER and other models are incomparable. The reason is that SHOPPER estimates the parameters across all categories, i.e., over the entire IRI dataset,  while our choice model estimation uses only the milk category.} The absolute value of the price coefficient estimated by our approach is significantly higher than that obtained by SHOPPER. We believe this reflects SHOPPER's limitation to account for price endogeneity. In terms of hit rates, we obtain comparable results, and our approach often outperforms SHOPPER. For example, in the case of using 20-dimensional vectors, our best model is Product2Vec (column 3), which increases the out-of-sample hit rate by 10\% compared with SHOPPER (14.1\% vs. 12.8\%).

\begin{table}
\centering
  \caption{Comparing Choice Models with SHOPPER}
  \label{tab:shopper-comparison-iv}
  \setlength{\tabcolsep}{2pt}
  \begin{tabular}{lcccccc}
    \toprule
    & \multicolumn{2}{c}{SHOPPER}  & \multicolumn{2}{c}{p2v}  & \multicolumn{2}{c}{rp2v}  \\ 
 \cmidrule(lr){2-3}\cmidrule(lr){4-5}\cmidrule(lr){6-7}
&\multicolumn{1}{c}{w/o}&\multicolumn{1}{c}{w}&\multicolumn{1}{c}{w/o}&\multicolumn{1}{c}{w}&\multicolumn{1}{c}{w/o}&\multicolumn{1}{c}{w}\\
    \midrule
Log price ($\beta$)         & -1.701        & -1.701        &  -3.166       &   -3.149      & -2.990                 &  -2.977                       \\
                       & --            & --           &     (0.136)         &     (0.136)             &     (0.132)         &     (0.131) \\  
In-sample   & \multirow{2}{*}{0.126}                      & \multirow{2}{*}{0.136}    &      \multirow{2}{*}{0.133}         &      \multirow{2}{*}{0.145}          &      \multirow{2}{*}{0.117}         &      \multirow{2}{*}{0.124} \\
hit rate \\
Out-of-sample   &   \multirow{2}{*}{0.119}                     & \multirow{2}{*}{0.128}                                   &      \multirow{2}{*}{0.141}         &      \multirow{2}{*}{0.138}           &      \multirow{2}{*}{0.110}         &     \multirow{2}{*}{0.132} \\
hit rate \\
    \bottomrule
\end{tabular}
\end{table}

\section{CONCLUSIONS}\label{sec:conclusions}

Our paper proposes Product2Vec, a method based on the representation learning algorithm Word2Vec, to understand product-level competition in large markets with millions of products. The Product2Vec model generates low-dimensional vectors to represent product attributes.

The comparative advantages of Product2Vec are multi-folds. First, our model is highly scalable and can be applied to millions of products because of the shallow model structure. Second, our model allows us to differentiate substitutes and complements using two metrics, exchangeability and complementarity. Third, by combining these vectors with random utility-based choice models, we can forecast demand more quickly and accurately. This is important for firms to make precise and timely predictions of future sales. Fourth, our model can estimate price elasticities more accurately by removing the influence of price from product vectors.

\bibliographystyle{ACM-Reference-Format}
\bibliography{reference}

\end{document}